\documentclass[letterpaper]{article}
\usepackage{amssymb}
\usepackage{aaai2026}
\usepackage{tablefootnote}
\usepackage{listings}
\usepackage{algpseudocode}
\usepackage{graphicx}

\def\xtest{x^{\textrm{test}}}
\def\xval{x^{\textrm{val}}}
\def\train{\textrm{train}}
\def\test{\textrm{test}}
\def\val{\textrm{val}}

\title{Resolving Predictive Multiplicity for the Rashomon Set}

\author{
    Parian Haghighat\textsuperscript{\rm 1},
    Hadis Anahideh\textsuperscript{\rm 1},
    Cynthia Rudin\textsuperscript{\rm 2}
}

\affiliations{
    \textsuperscript{\rm 1}University of Illinois Chicago, Chicago, IL, USA\\
    \textsuperscript{\rm 2}Duke University, Durham, NC, USA
}

\usepackage{times}
\usepackage{helvet}
\usepackage{courier}
\usepackage[hyphens]{url}
\usepackage{natbib}
\usepackage{subcaption}
\usepackage{xcolor}
\usepackage{amsmath} 
\usepackage{amsthm} 
\usepackage{multirow}
\usepackage{algorithm}

\usepackage{booktabs}

\begin{document}

\maketitle

\begin{abstract}
The existence of multiple, equally accurate models for a given predictive task leads to predictive multiplicity, where a ``Rashomon set'' of models achieve similar accuracy but diverge in their individual predictions. This inconsistency undermines trust in high-stakes applications where we want consistent predictions. We propose three approaches to reduce inconsistency among predictions for the members of the Rashomon set. The first approach is \textbf{outlier correction}. An outlier has a  label that none of the good models are capable of predicting correctly. Outliers can cause the Rashomon set to have high variance predictions in a local area, so fixing them can lower variance. Our second approach is \textbf{local patching}. In a local region around a test point, models may disagree with each other because some of them are biased. We can detect and fix such biases using a validation set, which also reduces multiplicity. Our third approach is \textbf{pairwise reconciliation}, where we find pairs of models that disagree on a region around the test point. We modify predictions that disagree, making them less biased. These three approaches can be used together or separately, and they each have distinct advantages. The reconciled predictions can then be distilled into a single interpretable model for real-world deployment. In experiments across multiple datasets, our methods reduce disagreement metrics while maintaining competitive accuracy.
\end{abstract}

\begin{links}
\link{Code}{https://github.com/parianh/ReconciledRashomon}
\end{links}

\section{Introduction}
The \textit{Rashomon Effect} is when a dataset admits a diverse set of models that perform about equally well \citep{McCullaghNelder1989,Breiman2001,RudinEtAlAmazing2024}. This gives rise to \textit{predictive multiplicity}, where models with similar global accuracy have divergent individual-level predictions. This has been observed in credit scoring, criminal sentencing, and medical diagnosis, where multiple equally accurate models disagree on individual predictions~\citep{Marx2020, Black2022, kobylinska2024exploration}. 

Model multiplicity has both practical advantages and disadvantages. On the positive side, the diversity of equally accurate models reflects the inherent flexibility of many learning problems and opens the door to aligning system behavior with crucial desiderata such as fairness, robustness, and interpretability \citep{Rudin2019,Black2022, RudinEtAlAmazing2024}. On the negative side, inconsistency in the predictions of equally good models potentially undermines trust in algorithmic decisions. Relying on only one good model risks reinforcing systemic blind spots and can leave decisions vulnerable to be challenged~\citep{kleinberg2021algorithmic, DAmour2020}.

In this work, we show how the Rashomon set, i.e., the set of good models for a given dataset, can be reconciled to agree on predictions in a way that does not harm accuracy. With each edit we make to a model in the Rashomon set, we lessen bias and improve local accuracy, which is why the model predictions iteratively converge toward agreement.

\begin{figure}[!t]
    \centering
    \includegraphics[width=\columnwidth]{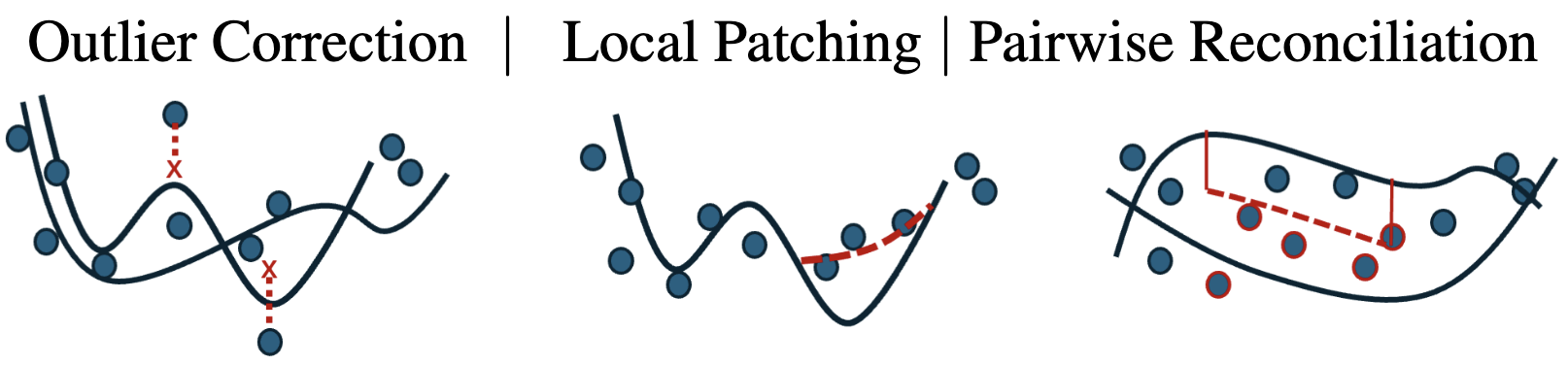}
    \caption{Illustration of approaches for regression.}
    \label{fig:approachesdiagram}
\end{figure}

We propose three approaches for reconciling predictions from the Rashomon set: \textbf{outlier correction}, \textbf{local patching}, and \textbf{pairwise reconciliation}, illustrated in Figure \ref{fig:approachesdiagram}; our algorithms are for classification, but it is easier to illustrate them for regression. These approaches can be used separately or together, and each has different benefits.  


\textbf{Outlier Correction (OC)}, as well as omitting outliers, is common as a data preprocessing step \citep{breunig2000lof,liu2012isolation, zhao2019pyod} to reduce the bias of models. Here, we use it for a different purpose: to reduce the variance of predictions among members of the Rashomon set. Models trained with outliers are generally worse and run the risk of overfitting. We identify outliers as points whose labels disagree with the predictions of models in the Rashomon set.

We consider both full OC (where we correct outliers in training and validation sets, then retrain) and a lighter OC variant that only corrects validation set outliers (without retraining).
As we show, outlier correction can improve predictive performance and stability, but it does not consistently reduce ensemble-wide disagreement on its own. We find that OC is most useful as a preprocessing step for subsequent reconciliation, in particular, correcting validation outlier labels experimentally provides cleaner data for our local patching method, discussed next.

In \textbf{local patching (LP)}, if a model exhibits systematic over- or under-prediction around a test point, we reduce this local bias by editing the model’s prediction at that point. Specifically, for each test point, we use a validation-set neighborhood to estimate directional bias for each model. We then patch predictions by adjusting them toward the neighborhood’s validation labels. This method empirically improves local calibration error and reduces local bias for models in the Rashomon set. 

Our third approach is \textbf{pairwise reconciliation (PR)}. This method is inspired by the theoretical reconciliation framework of \citet{Roth2023}, although our implementation addresses practical aspects that the theory does not. The general goal of reconciliation is to identify regions of systematic disagreement among accurate models and edit predictions to reduce inconsistency. 
Our approach differs from that of \citet{Roth2023} in important ways.
Our reconciliation decisions take advantage of held-out validation data, which we use to assess empirical suboptimality and verify that any proposed edits improve predictive performance. In addition, while \citet{Roth2023} perform reconciliation only by correcting one model to be similar to another in regions where they disagree, we make corrections by trading off predictive accuracy against resolving disagreement between the models. Finally, rather than reconciling model pairs in an arbitrary order, we prioritize the pairs of models exhibiting the largest disagreement, which improves computational efficiency and accelerates convergence in practice.

In our pairwise reconciliation approach,
(1) We iteratively identify the pair of models from the Rashomon set that disagree the most on predictions for the validation set;
(2) For each pair, we determine in which direction they most disagree (either the first model is larger more often or the second one is). The set of points on which the models disagree by at least $\epsilon$ in this direction is called the \textit{validation disagreement set}.
(3) We determine which of the two models is empirically suboptimal on the validation disagreement set -- i.e., has worse predictive performance on its validation labels -- and we say that model is \textit{falsified}.
(4) We then apply a correction to the falsified model's predictions on the validation disagreement set, where the correction magnitude is a linear combination of model error and disagreement between the models.
(5) We transfer the correction to test points where the same model pair disagrees in the same direction as in the \emph{validation disagreement set} used to compute that correction.

In this way, reconciliation edits predictions in regions of the test space where the disagreement structure is likely to persist. Empirically, pairwise reconciliation consistently and substantially reduces ensemble-wide disagreement across datasets, often driving variance, ambiguity, discrepancy, and disagreement-based metrics close to zero. However, reconciliation alone does not directly address cases where models share a common bias and therefore do not disagree—such as underfitting or systematic local miscalibration.
This is where local patching and outlier correction help. Local patching improves neighborhood agreement by correcting directional local bias, while outlier correction provides cleaner data for both local patching and reconciliation. We find that combining all three approaches yields robust reductions in disagreement while maintaining competitive predictive performance.

The reconciled predictions represent a consensus among good models; they are less locally biased than individual models and are more difficult to challenge than any single model. They have numerous possible options for downstream deployment; for instance, to support interpretability and accountability, we can distill the reconciled predictions into an interpretable meta-model that closely approximates the ensemble’s final behavior.
 
\section{Related Work}

\textbf{Predictive Multiplicity and the Rashomon Set.}
Throughout the history of statistics, scientists have noted the Rashomon Effect and the phenomenon of predictive multiplicity — when many models achieve near-identical accuracy, but produce conflicting predictions for the same points \citep{McCullaghNelder1989,Breiman2001,Rudin2019,  DAmour2020, Marx2020, Black2022, RudinEtAlAmazing2024}. The set of such well-performing but diverse models is called the \textit{Rashomon set} \citep{FisherRuDo19,SemenovaRuPa2022}, and recent work has focused on quantifying and characterizing the extent of disagreement among models within the Rashomon set. \citet{Marx2020} proposed a set of metrics to measure disagreement in classification tasks, which \citet{Watson2023b} extended to probabilistic predictions, highlighting that predictive variance is especially pronounced near decision boundaries and disproportionately affects individuals from minority groups. \citet{CokerRuKi21} defined tethered hacking intervals to measure the range of a statistic across the Rashomon set. If the statistic is a model prediction, the tethered hacking intervals will be prediction intervals that are robust to arbitrary choices made by scientists who might reasonably choose any model from the Rashomon set.

In parallel, advances in model enumeration techniques have enabled researchers to systematically explore the Rashomon set for specific model families. For example, complete enumerations have been developed for sparse decision trees \citep{Xin2022}, illustrating the sheer scale and structural diversity of high-performing models that can emerge from the same training data \citep{RudinEtAlAmazing2024}. 

Predictive multiplicity gives rise to what \citet{Black2022} term a ``crisis of justifiability.''  In high-stakes settings such as lending or criminal justice, if one model denies a loan while another—equally accurate—would approve it, the affected individual may have grounds to question the fairness and objectivity of the decision. Such inconsistencies at the individual level not only erode trust but also complicate contestability and recourse \citep{Long2024}.

\textbf{Reconciliation Strategies.} 
As discussed, \citet{Roth2023} propose a pairwise reconciliation framework that identifies regions of substantial disagreement between models and adjusts predictions to be closer on those disagreement regions. Our approach grounds decisions in held-out validation data and makes corrections by trading off predictive accuracy against reduced disagreement between models. Our approach is computationally efficient in practice, since it prioritizes the pairs with the most extreme disagreement rather than reconciling pairs in an arbitrary order.

\citet{Du2024reconciling} introduce a multi-calibration framework that reconciles predictions across groups by post-processing a base predictor (e.g., a score from an ensemble model) to be simultaneously calibrated on overlapping subpopulations. While this improves fairness metrics, it generally introduces additional bias that conflicts with our goals.

\section{Approaches}
We introduce our three approaches below. We are in a standard supervised binary classification setting, with i.i.d.\ points $\{x_i,y_i\}_{i=1}^n$, with $x\in\mathcal{X}$ and $y_i\in\{0,1\}$. We distinguish training points, validation points, and test points with superscripts or subscripts. The training set is $\mathcal{D}^{\mathrm{train}}$, the validation set is $\mathcal{D}^{\mathrm{val}}$, and the
test set is $\mathcal{D}^{\mathrm{test}}$.

We assume we have access to the Rashomon set, or a collection of good models. Complete Rashomon sets can be computed for sparse decision trees \citep{Xin2022} or linear models (which are ellipsoids around the least squares solution). A collection of good models can be constructed by running several different machine learning algorithms on the data; this is what we use in our experiments.  
The collection of models is denoted $f_m:\mathcal{X}\rightarrow [0,1]$ for $m=1..M$. The models predict the probability that $y=1\mid x$. We use $\mathrm{clip}_{[0,1]}(p)=\min\{1,\max\{0,p\}\}$ to ensure corrected probabilities remain in $[0,1]$.

We measure probabilistic prediction error using the \textbf{Brier loss}. 
For any finite dataset $\{(x_i,y_i)\}_{i=1}^n$ (e.g., the validation set $\mathcal D^{\mathrm{val}}$) with $y_i\in\{0,1\}$ and any index set 
$\mathcal I\subseteq\{1,\dots,n\}$, define the empirical Brier loss of a predictor $f$ on $\mathcal I$ as
\begin{equation}
\mathrm{BrierLoss}_{\mathcal I}(f)
:=\frac{1}{|\mathcal I|}\sum_{i\in\mathcal I}\bigl(f(x_i)-y_i\bigr)^2.
\label{eq:brierloss}
\end{equation}
When $\mathcal I=\{1,\dots,n_{\val}\}$, this corresponds to the standard validation Brier score. Our setup can be generalized to multiclass classification or regression, as well as to infinite sets of functions. 

\subsection{Outlier Correction}
Algorithm \ref{alg:correctingoutliers} shows how we correct outliers for classification, which improves label quality prior to reconciliation. Here, outliers are either true positives, where the average Rashomon set prediction is highly negative, or true negatives, where the average Rashomon set prediction is highly positive. For training outliers, we flip the label; this way, we can retrain the Rashomon set on the corrected labels afterwards if desired. For validation outliers, we replace the label with the mean predicted probability of being positive among the Rashomon set. 

One can create an analogous algorithm for regression, where an outlier would be a point that no or few good models can correctly predict. For instance, if the label is greater than all of the Rashomon set's predictions or lower than all of them, we would call it an outlier. We could also use a percentile (e.g., an outlier has a label above the 95th percentile or below the 5th percentile of predictions from the Rashomon set).

\begin{algorithm}[t]
\small
\caption{Outlier Correction (for classification)}
\label{alg:correctingoutliers}
\begin{algorithmic}[1]
\Require 
Training predictions $f_m(x^{\mathrm{train}}_i)$ for each training point $i$ and each model from the Rashomon set $m$, 
validation predictions $f_m(x^{\mathrm{val}}_i)$ for each validation point $i$ and each model from the Rashomon set $m$ (note that $i$ indexes both training and validation),  
labels $y^{\mathrm{train}}_i, y^{\mathrm{val}}_i$ for each $i$, 
ensemble size $M$, 
lower/upper thresholds for the estimated probability $(\tau_{\mathrm{low}}, \tau_{\mathrm{high}})$. 
\State Compute ensemble means for both training and validation points: $\hat f(x^{\train}_i) = \tfrac{1}{M}\sum_m f_{m}(x^{\train}_i)$ 
       and $\hat f(x^{\val}_i) = \tfrac{1}{M}\sum_m f_m(x^{\val}_i)$.
\State Identify outliers using the thresholds $(\tau_{\mathrm{low}}, \tau_{\mathrm{high}})$:
       $\mathcal{O} = \{ i : [(y^\mathrm{train\;or\;val}_i=1) \;\& \;(\hat f(x^{\mathrm{train\;or\;val}}_i) < \tau_{\mathrm{low}})] \;\text{or}\; [(y^\mathrm{train\;or\;val}_i=0) \;\& \;(\hat f(x^{\mathrm{train\;or\;val}}_i) > \tau_{\mathrm{high}})] \}$ 
       including points from both training and validation sets.
\State For each outlier $i$ in the training set, flip its label: $y^{\mathrm{train}}_i \leftarrow  1-y^{\mathrm{train}}_i$. For each outlier $i$ in the validation set, set $y^{\mathrm{val}}_i \leftarrow \hat f(x^{\mathrm{val}}_i)$.
\State If desired, recompute Rashomon set models using updated training labels $y^{\mathrm{train}}$.
\State \Return Updated labels, cleaned training data, and new Rashomon set.
\end{algorithmic}
\end{algorithm}

\subsection{Local Patching}

Algorithm \ref{alg:localpatching} describes local patching. Here, we find local regions where a model disagrees with the labels in its neighborhood, and we issue a patch to fix it. This directly improves the local accuracy of each model by identifying and correcting local biases, or ``blind spots.'' For each test instance $x^{\mathrm{test}}$, we first construct a context-specific neighborhood $N(x^{\mathrm{test}}) \subset \mathcal{D}^{\mathrm{val}}$ based on the radius of its $k$th nearest neighbor in the validation set. Within this neighborhood, each model $f_m$ is evaluated for systematic local bias. The residuals for model $f_m$ on validation neighbor $j$ are $R_m = \{ y^{\mathrm{val}}_j - f_m(x^{\mathrm{val}}_j) \}$; they are computed using validation points in the neighborhood $N(x^{\mathrm{test}})$. These residuals estimate the conditional error distribution $\varepsilon_m(x^{\mathrm{test}}) = \mathbb{E}_{Y|x^{\mathrm{test}}}[Y - f_m(x^{\mathrm{test}})]$ from the data. Under the assumption of local smoothness, the expectation of residuals in $N(x^{\text{test}})$ serves as a consistent estimator for the bias in $f_m$'s prediction at $x^{\text{test}}$.

To avoid overfitting to symmetric noise or canceling variance, we apply a one-sided bias test: we only compute correction terms when residuals exhibit directional skew above a threshold $\tau_{\text{bias}}$. Formally, the patch $\Delta_{m}$ estimates the conditional bias $\varepsilon_m(x^{\text{test}})$ when it is sufficiently large. The patched prediction $f'_m(x^{\text{test}}) = f_m(x^{\text{test}}) + \Delta_{m}$ thus serves as a locally calibrated output. 
\begin{equation}
\label{eq:localpatch}
\Delta_{m} =
\begin{cases}
  \frac{1}{|R_m^+|} \sum_{r \in R_m^+} r & \text{if } \frac{|R_m^+|}{|R_m|} > \tau_{\mathrm{bias}} \\
  \frac{1}{|R_m^-|} \sum_{r \in R_m^-} r & \text{if } \frac{|R_m^-|}{|R_m|} > \tau_{\mathrm{bias}} \\
  0 & \text{otherwise}
\end{cases}
\end{equation}
where $R_m^+ = \{r \in R_m \mid r > 0\}$ and $R_m^- = \{r \in R_m \mid r < 0\}$ are the sets of positive and negative residuals from $N(x^\mathrm{test})$, respectively. $\tau_{\mathrm{bias}} \in (0.5, 1]$ is a hyperparameter that controls sensitivity, and $|\cdot|$ is the count of elements in a set.
To avoid harmful corrections, after computing $\Delta_m$ from the validation neighborhood $N(x^{\test})$, we evaluate $\mathrm{BrierLoss}_{N(\xtest)}(f_m)$ on the validation neighbors in $N(x^{\test})$ before and after applying the (clipped) shift. We keep the patch at $x^{\test}$ only if this neighborhood Brier loss does not increase; otherwise, we set $\Delta_m=0$. The patched prediction is then $f'_m(x^{\test})=\mathrm{clip}_{[0,1]}\!\bigl(f_m(x^{\test})+\Delta_m\bigr)$, improving local calibration without worsening neighborhood Brier loss.

\begin{algorithm}[t]
\small
\caption{Local Patching}
\label{alg:localpatching}
\begin{algorithmic}[1]
\Require Validation set $\mathcal{D}^{\mathrm{val}}$, 
test set $\mathcal{D}^{\mathrm{test}}$, 
model predictions for the validation set $\{f_m(x^{\val}_i)\}_{m=1..M,i=1..n^{\val}}$, 
query point $\xtest$ and model predictions $\{f_m(\xtest)\}_{m=1}^M$,
neighbor parameters $(k, k_{\max})$, 
bias threshold $\tau_{\mathrm{bias}}$.
    \State Query the test point's $k_{\max}$ nearest neighbors within the validation set, KNN($\xtest$).
    \State Filter the KNN($\xtest$) to form neighborhood $N(\xtest)$.
    \For{each model $m = 1,\dots,M$}
        \State Compute residuals in $N(\xtest)$:
        \[
           R_m(\xtest) = \{ y^{\val}_j - f_m(\xval_j) : j \in N(\xtest) \}.
        \]
        \State Compute the fractions of positive and negative residuals, and compute patch $\Delta_{m}$ using Equation~\ref{eq:localpatch}.
        \Comment{keep the patch only if it does not worsen neighborhood Brier loss (otherwise set $\Delta_m\leftarrow 0$)}
        \If{$\Delta_m \neq 0$}
            \State $b_{\mathrm{before}} \leftarrow \frac{1}{|N(\xtest)|}\sum_{j\in N(\xtest)}\bigl(f_m(\xval_j)-y^{\val}_j\bigr)^2$
            \State $b_{\mathrm{after}} \leftarrow \frac{1}{|N(\xtest)|}\sum_{j\in N(\xtest)}\bigl(\mathrm{clip}_{[0,1]}(f_m(\xval_j)+\Delta_m)-y^{\val}_j\bigr)^2$
            \If{$b_{\mathrm{after}} > b_{\mathrm{before}}$}
                \State $\Delta_m \leftarrow 0$ \Comment{reject patch}
            \EndIf
        \EndIf
        \State $f'_m(\xtest) \leftarrow \mathrm{clip}_{[0,1]}\!\bigl(f_m(\xtest) + \Delta_m\bigr)$
\EndFor
\State \Return $f_m'(\xtest)$ for all $m$.
\end{algorithmic}
\end{algorithm}

\subsection{Pairwise Reconciliation}
\citet{Roth2023} points out that when two models substantially disagree on a region of the input space, at least one must be statistically suboptimal on that region and can be corrected. Repeating this process iteratively yields a low-bias consensus. 
In our approach, shown in Algorithm \ref{alg:pairwisereconciliation},
we iteratively identify pairs of models in the Rashomon set that exhibit the largest predictive disagreement on the validation set.  We quantify pairwise disagreement on validation using $\ell_1$ distance:
\[
D[m,j]=\frac{1}{n_{\val}}\sum_{i=1}^{n_{\val}}\bigl|f_m(\xval_i)-f_j(\xval_i)\bigr|,
\]
and at each iteration, we select the $B$ pairs $(m,j)$ with the largest $D[m,j]$ values.

We define the ensemble consensus on validation as the (fixed) initial mean prediction
\[
\hat f^{\mathrm{init}}(\xval_i)=\frac{1}{M}\sum_{m=1}^M f_m(\xval_i),
\]
computed once before iterations.

For each selected pair, we form the signed disagreement sets $\mathcal S^{>}$ and $\mathcal S^{<}$ (defined below) consisting of validation points where the two models differ by more than $\varepsilon$, and we select one of them (formalized below), denoted $\mathcal S$.

We then identify which of the two models performs worse on that set using validation labels; we say that this model is \textit{falsified} on $\mathcal S$. We then apply a correction to the falsified model’s predictions on this set, shifting them toward the observed labels as well as the ensemble’s consensus $\hat f^{\mathrm{init}}$.
We repeat this procedure until (i) the maximum pairwise disagreement falls below $\eta$, (ii) no further verified corrections are accepted (i.e., none satisfy the acceptance test), or (iii) the maximum number of iterations is reached.

By repeatedly reconciling the most disagreeing model pairs, we reduce disagreement across the ensemble while preserving accuracy.

There are several major differences between pairwise reconciliation and local patching. They have different goals: PR's goal is to encourage predictions of models to agree, whereas LP's goal is to reduce bias. Thus, LP considers individual models, whereas pairwise reconciliation considers pairs of models. LP applies a patch uniformly to a local region, whereas pairwise reconciliation applies corrections only at (or near) points where this same pair of models disagrees.

For each selected model pair $(m,j)$, we construct two \emph{signed disagreement sets} from the validation set:

\begin{align}
\mathcal{S}^{>} &= \Bigl\{ i\in\{1,\dots,n_{\val}\} : f_m(\xval_i)-f_j(\xval_i)>\varepsilon \Bigr\}, \label{eq:Sg}\\
\mathcal{S}^{<} &= \Bigl\{ i\in\{1,\dots,n_{\val}\} : f_j(\xval_i)-f_m(\xval_i)>\varepsilon \Bigr\}. \label{eq:Sl}
\end{align}
These sets capture points where one model exceeds the other by more than
the disagreement threshold $\varepsilon$.
We then choose the larger set and record its sign:
\begin{eqnarray*}
\mathcal S &\leftarrow& 
\arg\max_{\mathcal A\in\{\mathcal S^{>},\mathcal S^{<}\}} |\mathcal A|,\\
\sigma &\in& \{\texttt{>},\texttt{<}\}\text{ indicates whether }\mathcal S=\mathcal S^{>} \text{ or }\mathcal S^{<}.
\end{eqnarray*}
(If there is a tie, we take $\mathcal S=\mathcal S^{>}$, matching Algorithm~\ref{alg:pairwisereconciliation}.)

If $|\mathcal S|<\alpha$, we skip this pair. Otherwise, let $k$ denote the falsified model, i.e., the one with higher $\mathrm{BrierLoss}_{\mathcal S}$ over $\mathcal S$ (see Eq.~\eqref{eq:brierloss}).

We then compute the correction $z^*$ by minimizing the empirical convex objective over the selected disagreement index set $\mathcal S$:
\[
\begin{aligned}
F(z)
&= \lambda\,\frac{1}{|\mathcal S|}\sum_{i\in\mathcal S}\Bigl(f_k(\xval_i)+z-y^{\val}_i\Bigr)^2 \\
&\quad+\; (1-\lambda)\,\frac{1}{|\mathcal S|}\sum_{i\in\mathcal S}\Bigl(f_k(\xval_i)+z-\hat f^{\mathrm{init}}(\xval_i)\Bigr)^2
\end{aligned}
\]
whose closed-form minimizer is
\begin{equation}
z^* \;=\; \lambda\,(\overline{y}_{\mathcal{S}}^{\mathrm{val}} - \overline{f}_{k,\mathcal{S}}^{\mathrm{val}}) + (1-\lambda)\,(\overline{\hat{f}}_{\mathcal{S}}^{\mathrm{init}} - \overline{f}_{k,\mathcal{S}}^{\mathrm{val}})
\label{eq:zstar_closed_form}
\end{equation}
where $\overline{y}_{\mathcal{S}}^{\mathrm{val}}$, $\overline{f}_{k,\mathcal{S}}^{\mathrm{val}}$, and $\overline{\hat{f}}_{\mathcal{S}}^{\mathrm{init}}$ denote the averages of $y_i^{\val}$, $f_k(\xval_i)$, and $\hat f^{\mathrm{init}}(\xval_i)$ over $i\in\mathcal S$, respectively. Here, $z^*$ is a weighted correction that shifts the falsified model toward both the true labels and the ensemble’s predictive consensus.

A correction is accepted only if it reduces the Brier loss of the falsified model $k$ on $\mathcal S$ by at least $\delta$.
If accepted, we export the same mean shift $z^*$ to the corresponding signed disagreement region on the test set. To define this region, we construct the signed disagreement sets on the test set analogously:
\begin{align*}
\mathcal S^{>}_{\test}&=\Bigl\{ i\in\{1,\dots,n_{\test}\} : f_m(\xtest_i)-f_j(\xtest_i)>\varepsilon \Bigr\},\\
\mathcal S^{<}_{\test}&=\Bigl\{ i\in\{1,\dots,n_{\test}\} : f_j(\xtest_i)-f_m(\xtest_i)>\varepsilon \Bigr\}.
\end{align*}
When $\mathcal S=\mathcal S^{>}$ (resp.\ $\mathcal S=\mathcal S^{<}$) on validation, we export the shift to
$\mathcal S^{\test}=\mathcal S^{>}_{\test}$ (resp.\ $\mathcal S^{\test}=\mathcal S^{<}_{\test}$).

For classification tasks, corrected predictions are clipped to $[0,1]$ to preserve probabilistic validity.  
Iterations continue until no further verified corrections can be applied, yielding a reconciled ensemble with reduced internal disagreement and improved calibration consistency. Algorithm~\ref{alg:pairwisereconciliation} summarizes the full procedure.
\section{Experimental Setup}

\subsection{Datasets}
We evaluate four high-stakes \emph{classification} tasks (Table~\ref{tab:datasets}): Adult~\citep{adult_2}, COMPAS~\citep{larson2016compas}, and two Folktables tasks, Folk Mobility and Folk Travel~\citep{ding2021}. All datasets use a 60/20/20 train/validation/test split and results are averaged over 10 random seeds.

\begin{table}[ht]
\centering
\scriptsize
\begin{tabular}{lccc}
\toprule
\textbf{Dataset} & \textbf{Size} & \textbf{Features} & \textbf{Target} \\
\midrule
\textit{Adult} (UCI) & 32,561 & 97 & Income $>$ 50K \\
\textit{COMPAS} & 6,172 & 351 & 2-year Recidivism \\
\textit{Folk Mobility} & 34,491 & 96 & Mobility \\
\textit{Folk Travel} & 88,071 & 71 & Travel Time \\
\bottomrule
\end{tabular}
\caption{Summary of classification datasets used in experiments. Feature counts reflect processed features, where categorical features were encoded as binary features.}
\label{tab:datasets}
\end{table}

\subsection{Evaluation Metrics}
\label{sec:metrics}
We evaluate methods according to: \textbf{(i) predictive performance}, \textbf{(ii) predictive multiplicity}, and \textbf{(iii) neighborhood agreement}. Results are reported as mean $\pm$ one standard deviation over 10 random seeds.

\begin{algorithm}[p]
\footnotesize
\caption{Pairwise Reconciliation}
\label{alg:pairwisereconciliation}
\begin{algorithmic}[1]
\Require 
Model predictions $\{f_m(\xval_i)\}_{m=1..M,\ i=1..n_{\val}}$,
$\{f_m(\xtest_i)\}_{m=1..M,\ i=1..n_{\test}}$,
labels $\{y^{\val}_i\}_{i=1}^{n_{\val}}$,
disagreement threshold $\varepsilon$, 
batch size $B$,
minimum region mass $\alpha$,
objective weight $\lambda$, 
acceptance tolerance $\delta$,
disagreement tolerance $\eta$, 
maximum iterations $T_{\max}$.
\State Compute mean prediction on validation:
$\hat f^{\mathrm{init}}(\xval_i)\leftarrow \tfrac{1}{M}\sum_{m=1}^M f_m(\xval_i)$ for all $i\in\{1,\dots,n_{\val}\}$.
\State Initialize $t\!\leftarrow\!0$.
\While{$t < T_{\max}$}
  \Comment{Compute pairwise disagreements on validation}
  \ForAll{ $m$ and $j$ such that $m<j$}
    \State $D[m,j]\leftarrow \tfrac{1}{n_{\val}}\sum_{i=1}^{n_{\val}} \big| f_m(\xval_i)-f_j(\xval_i)\big|$.
  \EndFor
  \If{$\max_{m,j: m<j} D[m,j] < \eta$}
    \State \textbf{break out of while loop}
  \EndIf
  \State Select top-$B$ disagreeing pairs $\mathcal P_t$ according to $D[m,j]$.
  \State $\texttt{acceptedAtLeastOnce}\leftarrow \texttt{false}$.
  \ForAll{$(m,j)\in\mathcal P_t$}
    \State $\mathcal S^{>}\leftarrow\{i: f_m(\xval_i)-f_j(\xval_i)>\varepsilon\}$.
    \State $\mathcal S^{<}\leftarrow\{i: f_j(\xval_i)-f_m(\xval_i)>\varepsilon\}$.
    \Comment{Define validation disagreement set $\mathcal{S}$}
    \If{$|\mathcal S^{>}|\ge|\mathcal S^{<}|$}
      \State $\mathcal S\leftarrow \mathcal S^{>}$ and $\sigma\leftarrow \texttt{>}$.
    \Else
      \State $\mathcal S\leftarrow \mathcal S^{<}$ and $\sigma\leftarrow \texttt{<}$.
    \EndIf
    \If{$|\mathcal S| < \alpha$}
      \State \textbf{skip to the next pair in $\mathcal P_t$}.
    \Else
        \Comment{Determine which model is falsified} 
        \State $k\leftarrow \arg\max_{u\in\{m,j\}} \mathrm{BrierLoss}_{\mathcal S}(f_u)$.
        \State $\overline f^{\val}_{k,\mathcal S}\leftarrow \tfrac{1}{|\mathcal S|}\sum_{i\in\mathcal S} f_k(\xval_i)$,
        \State $\overline y^{\val}_{\mathcal S}\leftarrow \tfrac{1}{|\mathcal S|}\sum_{i\in\mathcal S} y^{\val}_i$,
        \State $\overline{\hat f}^{\mathrm{init}}_{\mathcal S}\leftarrow \tfrac{1}{|\mathcal S|}\sum_{i\in\mathcal S} \hat f^{\mathrm{init}}(\xval_i)$.
        \State Compute closed-form correction $z^*$ from Eq.~\eqref{eq:zstar_closed_form}.
        \State For all $i\in\mathcal S$, set $f_k'(\xval_i)\leftarrow \mathrm{clip}_{[0,1]}\!\bigl(f_k(\xval_i)+z^*\bigr)$.
        \If{$\mathrm{BrierLoss}_{\mathcal S}(f_k') < \mathrm{BrierLoss}_{\mathcal S}(f_k)-\delta$}
            \Comment{Transfer correction to test set}
            \State Accept: for all $i\in\mathcal S$, set $f_k(\xval_i)\leftarrow f_k'(\xval_i)$.
        \If{$\sigma=\texttt{>}$}
            \State $\mathcal S^{\test}\leftarrow\{i: f_m(\xtest_i)-f_j(\xtest_i)>\varepsilon\}$.
        \Else
            \State $\mathcal S^{\test}\leftarrow\{i: f_j(\xtest_i)-f_m(\xtest_i)>\varepsilon\}$.
        \EndIf
        \State $\texttt{acceptedAtLeastOnce}\leftarrow \texttt{true}$.
        \State For all $i\in\mathcal S^{\test}$, set
        \State $f_k(\xtest_i)\leftarrow \mathrm{clip}_{[0,1]}\!\bigl(f_k(\xtest_i)+z^*\bigr)$.

        \EndIf
    \EndIf
  \EndFor
  \If{\textbf{not} \texttt{acceptedAtLeastOnce}}
    \State \textbf{break} \Comment{no accepted corrections; models already sufficiently agree}
  \EndIf
  \State $t\leftarrow t+1$.
\EndWhile
\State \Return reconciled $\{f_m(\xval_i)\}_{m=1..M,\ i=1..n_{\val}}$ and reconciled $\{f_m(\xtest_i)\}_{m=1..M,\ i=1..n_{\test}}$.
\end{algorithmic}
\end{algorithm}
\paragraph{Predictive Performance.}
We report test-set metrics to ensure reconciliation does not degrade predictive quality:
\textbf{Accuracy} (thresholding probabilities at $0.5$) and the \textbf{Brier score} (mean squared error between predicted probabilities and binary labels).
For each method, we compute these metrics using its ensemble prediction $\hat f(x^{\mathrm{test}})\in[0,1]$ on $\mathcal{D}^{\mathrm{test}}$.

\paragraph{Predictive Multiplicity.}
We report four disagreement-based metrics computed from the set of model predictions
$\{f_m(x^{\mathrm{test}})\}_{m=1}^M$ on the test set. We define:
(i) \textbf{Variance}, the mean over $x^{\mathrm{test}}$ of $\mathrm{Var}_{m}\!\left[f_m(x^{\mathrm{test}})\right]$;
(ii) \textbf{Ambiguity}, the mean over $x^{\mathrm{test}}$ of $\max_m f_m(x^{\mathrm{test}})-\min_m f_m(x^{\mathrm{test}})$;
(iii) \textbf{Discrepancy}, the maximum over model pairs $(m,j)$ of the mean absolute difference
$\tfrac{1}{n_{\mathrm{test}}}\sum_{x^{\mathrm{test}}\in\mathcal{D}^{\mathrm{test}}} \left| f_m(x^{\mathrm{test}})-f_j(x^{\mathrm{test}}) \right|$; and
(iv) \textbf{Disagreement Rate}, the mean over model pairs $(m,j)$ of the fraction of test points for which
$\left| f_m(x^{\mathrm{test}})-f_j(x^{\mathrm{test}}) \right|>\varepsilon$, with $\varepsilon=0.05$.

\paragraph{Neighborhood Agreement.}
We report \textbf{Local Conditional Absolute Error (LCAE)}, computed using validation neighborhoods, as a kNN-based proxy for local conditional error. We interpret lower LCAE as better \textbf{neighborhood agreement}, i.e., closer alignment between a test prediction and outcomes among nearby validation points.
For each test point $x_i^{\mathrm{test}}$, let $\mathcal{N}_{30}(x_i^{\mathrm{test}})\subset\mathcal{D}^{\mathrm{val}}$ denote its $k=30$ nearest neighbors in the validation set (in the preprocessed feature space). We define
\begin{equation}
\small
\mathrm{LCAE}_{30}(x_i^{\mathrm{test}})
=\frac{1}{|\mathcal{N}_{30}(x_i^{\mathrm{test}})|}
\sum_{x_j^{\mathrm{val}}\in \mathcal{N}_{30}(x_i^{\mathrm{test}})}
\left| \hat f(x_i^{\mathrm{test}})-y^{\mathrm{val}}_j \right|,
\end{equation}
where $\hat f(x_i^{\mathrm{test}})$ is the method's final ensemble prediction at $x_i^{\mathrm{test}}$ and $y^{\mathrm{val}}_j\in\{0,1\}$ is the validation label for neighbor $x_j^{\mathrm{val}}$.
We report the average over the test set:
\[
\mathrm{LCAE}
=\frac{1}{n_{\mathrm{test}}}\sum_{i=1}^{n_{\mathrm{test}}} \mathrm{LCAE}_{30}(x_i^{\mathrm{test}}),
\]
where lower values indicate better neighborhood agreement.

\subsection{Experimental Protocol}
\label{sec:exp_protocol}

To approximate a Rashomon set, we train a diverse pool of probabilistic classifiers in scikit-learn (logistic regression, random forests, extra trees, gradient boosting, and multi-layer perceptrons) using fixed hyperparameter grids. We evaluate candidates on the validation set and retain the top $M=25$ models with the lowest validation Brier score. 

Each dataset is split into 60\%/20\%/20\% train/validation/test with stratification on the target label. The validation split is used (i) to define neighborhoods for neighborhood agreement (LCAE) and local patching, (ii) to select outlier-correction thresholds, (iii) to evaluate candidates for Rashomon set selection, and (iv) to identify disagreement regions and verify pairwise corrections.
All results are reported as mean $\pm$ one standard deviation over 10 random seeds.

\subsection{Baselines} 
These baselines produce a single predictor from the Rashomon-set predictions $\{f_m(x)\}_{m=1}^M$:
\begin{itemize}
    \item \textbf{Soft Voting} (ensemble mean): $\hat f(x)=\tfrac{1}{M}\sum_{m=1}^M f_m(x)$.
    \item \textbf{Majority Voting}: binarize each model at $0.5$, average votes, and interpret the vote fraction as a probability.    
    \item \textbf{Best Single Model}: choose the model in the Rashomon set with the highest validation accuracy (breaking ties by lower validation Brier score) and use it for all test points.
    \item \textbf{Random Selection}: for each test point, sample one model index uniformly at random and use its prediction.
\end{itemize}
Since these methods do not modify the underlying model predictions $\{f_m\}$, the ensemble-wide disagreement metrics (Var/Amb/Disc/Disag) are computed on the same underlying set $\{f_m\}$ and are thus identical across aggregation baselines.

\subsection{Our approaches}

\paragraph{Outlier Correction (OC).}
We evaluate OC, as in Algorithm~\ref{alg:correctingoutliers}.
Outliers are identified using the Rashomon mean prediction and thresholds $(\tau_{\mathrm{low}},\tau_{\mathrm{high}})$ that come from hyperparameter tuning.
Training outlier labels are flipped and the ensemble is retrained; validation outlier labels are replaced with the Rashomon mean prediction.

\paragraph{Local Patching (LP).}
To evaluate LP, for each test point $x^{\test}$, we form a neighborhood $N(x^{\test})\subset \mathcal{D}^{\val}$ via $k$-nearest neighbors in the preprocessed feature space.
Each model is locally corrected via the one-sided residual rule in Eq.~\eqref{eq:localpatch}, with threshold $\tau_{\mathrm{bias}}$, chosen via hyperparameter tuning on the validation set.

\paragraph{Pairwise Reconciliation (PR).} 
To evaluate PR, we run Algorithm~\ref{alg:pairwisereconciliation} on each trained Rashomon set, using the validation split to identify the top-$B$ most-disagreeing pairs, form disagreement regions, and accept corrections only when they reduce validation Brier loss by at least $\delta$. Hyperparameters $(\varepsilon, B, \alpha, \lambda, \delta, \eta, T_{\max})$ are selected via validation tuning, and we report performance, multiplicity, and neighborhood agreement (LCAE) on the resulting reconciled test predictions.

\paragraph{Combinations.}
We evaluate combinations of our methods, specifically, OC+LP, OC+PR, PR+LP, and OC+PR+LP. 

\section{Results}
\label{sec:results}

Tables~\ref{tab:adultTable}--\ref{tab:FolkTTable} summarize the relative performance of all methods across the four datasets in Table~\ref{tab:datasets}, reporting mean percent change ($\Delta\%$) relative to soft voting over 10 random seeds. Improvements are shown in green and degradations in pink.

We evaluate methods along three dimensions: predictive accuracy, neighborhood agreement (LCAE@30), and predictive multiplicity (Variance, Ambiguity, Discrepancy, and Disagreement Rate). The results highlight that predictive multiplicity is multi-dimensional.

\textbf{Predictive Performance.} Across all datasets, the combined methods preserve and often improve predictive accuracy relative to soft voting. In particular, PR+LP and OC+PR+LP achieve accuracy that is comparable to or better than PR alone in most settings. For example, OC+PR+LP improves accuracy on Adult, Folk Mobility, and Folk Travel while substantially reducing disagreement. On COMPAS, accuracy changes remain small across all reconciliation-based variants, indicating that disagreement reduction does not come at the cost of substantial performance degradation.

Local Patching alone tends to incur modest accuracy losses, which is not intuitive and could potentially arise from overfitting based on our parameter choices. We used only 5 nearest neighbors to define the neighborhood, which is likely too small, and could be better chosen in future work. When LP is paired with reconciliation in our experiments, the losses are largely mitigated. Aggregation baselines such as majority voting and best single model show inconsistent accuracy and no systematic improvements.

\textbf{Reduction of Predictive Multiplicity.} 
Predictive multiplicity is most effectively reduced through reconciliation-based methods, but the results show that the most robust reductions arise from combinations rather than reconciliation alone. While PR alone drives dramatic reductions in disagreement (e.g., -99.67\% on Adult and -99.77\% on COMPAS), PR+LP and OC+PR+LP achieve similarly large reductions across all disagreement metrics while offering additional benefits in local agreement. PR+LP and OC+PR+LP achieve disagreement reductions comparable to PR alone, often exceeding -87\% across datasets.

In contrast, Local Patching alone produces only moderate reductions in disagreement, and Outlier Correction without reconciliation does not consistently reduce ensemble-wide multiplicity. Aggregation baselines leave disagreement metrics unchanged, underscoring that modifying predictions rather than merely aggregating them substantially resolves predictive multiplicity.

\textbf{Neighborhood Agreement (LCAE@30).}
Neighborhood agreement reveals the clearest advantage of combining methods. Local Patching is the primary driver of LCAE@30 improvements, consistently reducing neighborhood error across all datasets. However, LP alone does not eliminate disagreement and can negatively affect accuracy. When LP is combined with reconciliation, these limitations are addressed.
PR+LP consistently achieves the largest reductions in LCAE@30 among reconciliation-based methods, such as -13.96\% on Adult, -10.39\% on Folk Mobility, and -6.00\% on Folk Travel. OC+PR+LP also yields substantial improvements in neighborhood agreement while maintaining strong disagreement reduction and stable accuracy.

Pairwise reconciliation alone yields only modest LCAE improvements, indicating that reducing disagreement does not necessarily correct shared local bias. This distinction highlights the complementary roles of reconciliation and local patching: reconciliation aligns models with one another, while patching aligns predictions with outcomes.

\textbf{Combined Effects and Practical Trade-offs.} 
Taken together, the results show that no single mechanism is sufficient to address all aspects of predictive reliability. Pairwise reconciliation eliminates ensemble-wide disagreement, but it is insufficient for improving neighborhood agreement on its own. Local patching corrects local bias, and outlier correction stabilizes downstream corrections by improving label quality.

The strongest overall trade-offs are consistently obtained by PR+LP and OC+PR+LP, achieving low disagreement, improved neighborhood agreement, and competitive accuracy.
\begin{table}[!t]
\centering
\scriptsize
\begin{tabular}{l@{}r@{}r@{}r@{}r@{}r@{}r}
\toprule
\textbf{Method}
& $\Delta\text{Acc}\uparrow$ 
& $\Delta\text{LCAE}\downarrow$ 
& $\Delta\text{Var}\downarrow$ 
& $\Delta\text{Amb}\downarrow$ 
& $\Delta\text{Disc}\downarrow$ 
& $\Delta\text{Disag}\downarrow$ \\
\midrule
Soft Voting & +0.00\% & +0.00\% & +0.00\% & +0.00\% & +0.00\% & +0.00\% \\
Random Selection & \textcolor{magenta}{-0.95\%} & \textcolor{magenta}{+0.04\%} & +0.00\% & +0.00\% & +0.00\% & +0.00\% \\
Majority Voting & \textcolor{teal}{+0.07\%} & \textcolor{teal}{-24.15\%} & +0.00\% & +0.00\% & +0.00\% & +0.00\% \\
Best Single Model & \textcolor{teal}{+0.37\%} & \textcolor{teal}{-7.17\%} & +0.00\% & +0.00\% & +0.00\% & +0.00\% \\
\hline
LP & \textcolor{magenta}{-0.49\%} & \textcolor{teal}{-13.88\%} & \textcolor{teal}{-16.98\%} & \textcolor{teal}{-27.80\%} & \textcolor{teal}{-33.73\%} & \textcolor{teal}{-39.95\%} \\
OC + LP & +0.00\% & \textcolor{teal}{-12.08\%} & \textcolor{teal}{-7.55\%} & \textcolor{teal}{-17.45\%} & \textcolor{teal}{-24.34\%} & \textcolor{teal}{-27.49\%} \\
PR & \textcolor{teal}{+0.28\%} & \textcolor{teal}{-5.08\%} & \textcolor{teal}{-98.11\%} & \textcolor{teal}{-84.59\%} & \textcolor{teal}{-90.14\%} & \textcolor{teal}{-99.67\%} \\
OC + PR & \textcolor{teal}{+0.32\%} & \textcolor{teal}{-5.00\%} & \textcolor{teal}{-83.02\%} & \textcolor{teal}{-55.78\%} & \textcolor{teal}{-70.96\%} & \textcolor{teal}{-67.20\%} \\
PR + LP & \textcolor{magenta}{-0.35\%} & \textcolor{teal}{-13.96\%} & \textcolor{teal}{-96.23\%} & \textcolor{teal}{-86.91\%} & \textcolor{teal}{-92.44\%} & \textcolor{teal}{-98.66\%} \\
OC + PR + LP & \textcolor{teal}{+0.25\%} & \textcolor{teal}{-11.46\%} & \textcolor{teal}{-98.11\%} & \textcolor{teal}{-87.52\%} & \textcolor{teal}{-92.68\%} & \textcolor{teal}{-98.77\%} \\
\bottomrule
\end{tabular}
\caption{Adult dataset.}
\label{tab:adultTable}
\end{table}

\begin{table}[!t]
\centering
\scriptsize
\begin{tabular}{l@{}r@{}r@{}r@{}r@{}r@{}r}
\toprule
\textbf{Method}
& $\Delta\text{Acc}\uparrow$ 
& $\Delta\text{LCAE}\downarrow$ 
& $\Delta\text{Var}\downarrow$ 
& $\Delta\text{Amb}\downarrow$ 
& $\Delta\text{Disc}\downarrow$ 
& $\Delta\text{Disag}\downarrow$ \\
\midrule
Soft Voting & +0.00\% & +0.00\% & +0.00\% & +0.00\% & +0.00\% & +0.00\% \\
Random Selection & \textcolor{magenta}{-3.29\%} & +0.00\% & +0.00\% & +0.00\% & +0.00\% & +0.00\% \\
Majority Voting & \textcolor{teal}{+0.12\%} & \textcolor{teal}{-15.12\%} & +0.00\% & +0.00\% & +0.00\% & +0.00\% \\
Best Single Model & \textcolor{teal}{+0.43\%} & \textcolor{teal}{-2.81\%} & +0.00\% & +0.00\% & +0.00\% & +0.00\% \\
\hline
LP & \textcolor{magenta}{-1.82\%} & \textcolor{teal}{-7.96\%} & \textcolor{teal}{-3.06\%} & \textcolor{teal}{-12.04\%} & \textcolor{teal}{-18.65\%} & \textcolor{teal}{-18.93\%} \\
OC + LP & \textcolor{teal}{+0.10\%} & \textcolor{teal}{-4.99\%} & \textcolor{magenta}{+5.10\%} & +0.00\% & \textcolor{teal}{-2.25\%} & \textcolor{teal}{-4.95\%} \\
PR & \textcolor{magenta}{-0.73\%} & \textcolor{teal}{-2.11\%} & \textcolor{teal}{-98.98\%} & \textcolor{teal}{-89.62\%} & \textcolor{teal}{-93.69\%} & \textcolor{teal}{-99.77\%} \\
OC + PR & \textcolor{teal}{+0.26\%} & \textcolor{teal}{-2.13\%} & \textcolor{teal}{-82.65\%} & \textcolor{teal}{-56.28\%} & \textcolor{teal}{-68.02\%} & \textcolor{teal}{-57.10\%} \\
PR + LP & \textcolor{magenta}{-2.27\%} & \textcolor{teal}{-7.18\%} & \textcolor{teal}{-96.94\%} & \textcolor{teal}{-88.31\%} & \textcolor{teal}{-92.59\%} & \textcolor{teal}{-98.46\%} \\
OC + PR + LP & \textcolor{magenta}{-0.74\%} & \textcolor{teal}{-4.48\%} & \textcolor{teal}{-97.96\%} & \textcolor{teal}{-88.83\%} & \textcolor{teal}{-92.87\%} & \textcolor{teal}{-99.03\%} \\
\bottomrule
\end{tabular}
\caption{COMPAS dataset.\label{tab:CompasTable}}
\end{table}

\begin{table}[!t]
\centering
\scriptsize
\begin{tabular}{l@{}r@{}r@{}r@{}r@{}r@{}r}
\toprule
\textbf{Method}
& $\Delta\text{Acc}\uparrow$ 
& $\Delta\text{LCAE}\downarrow$ 
& $\Delta\text{Var}\downarrow$ 
& $\Delta\text{Amb}\downarrow$ 
& $\Delta\text{Disc}\downarrow$ 
& $\Delta\text{Disag}\downarrow$ \\
\midrule
Soft Voting & +0.00\% & +0.00\% & +0.00\% & +0.00\% & +0.00\% & +0.00\% \\
Random Selection & \textcolor{magenta}{-1.17\%} & \textcolor{teal}{-0.03\%} & +0.00\% & +0.00\% & +0.00\% & +0.00\% \\
Majority Voting & \textcolor{magenta}{-0.76\%} & \textcolor{teal}{-28.82\%} & +0.00\% & +0.00\% & +0.00\% & +0.00\% \\
Best Single Model & \textcolor{magenta}{-0.06\%} & \textcolor{teal}{-1.19\%} & +0.00\% & +0.00\% & +0.00\% & +0.00\% \\
\hline
LP & \textcolor{magenta}{-0.80\%} & \textcolor{teal}{-10.06\%} & \textcolor{magenta}{+5.26\%} & \textcolor{teal}{-10.40\%} & \textcolor{teal}{-12.25\%} & \textcolor{teal}{-20.82\%} \\
OC + LP & \textcolor{magenta}{-0.07\%} & \textcolor{teal}{-8.04\%} & \textcolor{magenta}{+7.89\%} & \textcolor{teal}{-3.34\%} & \textcolor{teal}{-6.76\%} & \textcolor{teal}{-10.18\%} \\
PR & \textcolor{teal}{+0.99\%} & \textcolor{teal}{-1.28\%} & \textcolor{teal}{-92.11\%} & \textcolor{teal}{-69.67\%} & \textcolor{teal}{-78.71\%} & \textcolor{teal}{-90.53\%} \\
OC + PR & \textcolor{teal}{+0.88\%} & \textcolor{teal}{-2.48\%} & \textcolor{teal}{-71.05\%} & \textcolor{teal}{-48.78\%} & \textcolor{teal}{-63.00\%} & \textcolor{teal}{-57.67\%} \\
PR + LP & \textcolor{magenta}{-0.08\%} & \textcolor{teal}{-10.39\%} & \textcolor{teal}{-81.58\%} & \textcolor{teal}{-69.91\%} & \textcolor{teal}{-79.58\%} & \textcolor{teal}{-89.26\%} \\
OC + PR + LP & \textcolor{teal}{+0.91\%} & \textcolor{teal}{-7.85\%} & \textcolor{teal}{-86.84\%} & \textcolor{teal}{-68.91\%} & \textcolor{teal}{-78.32\%} & \textcolor{teal}{-87.69\%} \\
\bottomrule
\end{tabular}
\caption{Folk Mobility dataset.\label{tab:FolkMTable}}
\end{table}

\begin{table}[!t]
\centering
\scriptsize
\begin{tabular}{l@{}r@{}r@{}r@{}r@{}r@{}r}
\toprule
\textbf{Method}
& $\Delta\text{Acc}\uparrow$ 
& $\Delta\text{LCAE}\downarrow$ 
& $\Delta\text{Var}\downarrow$ 
& $\Delta\text{Amb}\downarrow$ 
& $\Delta\text{Disc}\downarrow$ 
& $\Delta\text{Disag}\downarrow$ \\
\midrule
Soft Voting & +0.00\% & +0.00\% & +0.00\% & +0.00\% & +0.00\% & +0.00\% \\
Random Selection & \textcolor{magenta}{-4.16\%} & +0.00\% & +0.00\% & +0.00\% & +0.00\% & +0.00\% \\
Majority Voting & \textcolor{magenta}{-0.54\%} & \textcolor{teal}{-7.55\%} & +0.00\% & +0.00\% & +0.00\% & +0.00\% \\
Best Single Model & \textcolor{teal}{+2.61\%} & \textcolor{teal}{-1.99\%} & +0.00\% & +0.00\% & +0.00\% & +0.00\% \\
\hline
LP & \textcolor{magenta}{-4.19\%} & \textcolor{teal}{-6.15\%} & \textcolor{magenta}{+8.93\%} & \textcolor{teal}{-11.07\%} & \textcolor{teal}{-13.57\%} & \textcolor{teal}{-21.86\%} \\
OC + LP & \textcolor{magenta}{-0.14\%} & \textcolor{teal}{-2.28\%} & \textcolor{magenta}{+5.36\%} & \textcolor{teal}{-0.77\%} & \textcolor{teal}{-1.76\%} & \textcolor{teal}{-5.04\%} \\
PR & \textcolor{teal}{+2.58\%} & \textcolor{teal}{-1.62\%} & \textcolor{teal}{-94.64\%} & \textcolor{teal}{-76.93\%} & \textcolor{teal}{-81.07\%} & \textcolor{teal}{-96.06\%} \\
OC + PR & \textcolor{teal}{+2.36\%} & \textcolor{teal}{-1.45\%} & \textcolor{teal}{-80.36\%} & \textcolor{teal}{-54.75\%} & \textcolor{teal}{-63.39\%} & \textcolor{teal}{-57.08\%} \\
PR + LP & \textcolor{magenta}{-2.34\%} & \textcolor{teal}{-6.00\%} & \textcolor{teal}{-85.71\%} & \textcolor{teal}{-75.78\%} & \textcolor{teal}{-81.29\%} & \textcolor{teal}{-93.94\%} \\
OC + PR + LP & \textcolor{teal}{+2.42\%} & \textcolor{teal}{-2.85\%} & \textcolor{teal}{-92.86\%} & \textcolor{teal}{-76.62\%} & \textcolor{teal}{-81.07\%} & \textcolor{teal}{-95.59\%} \\
\bottomrule
\end{tabular}
\caption{Folk Travel dataset.\label{tab:FolkTTable}}
\end{table}
\bibliography{aaai2026}
\newpage
\appendix
\section{Appendix}

\subsection{Accuracy-Disagreement Trade-offs}
\begin{figure}[H]
  \centering
  \includegraphics[width=\columnwidth]{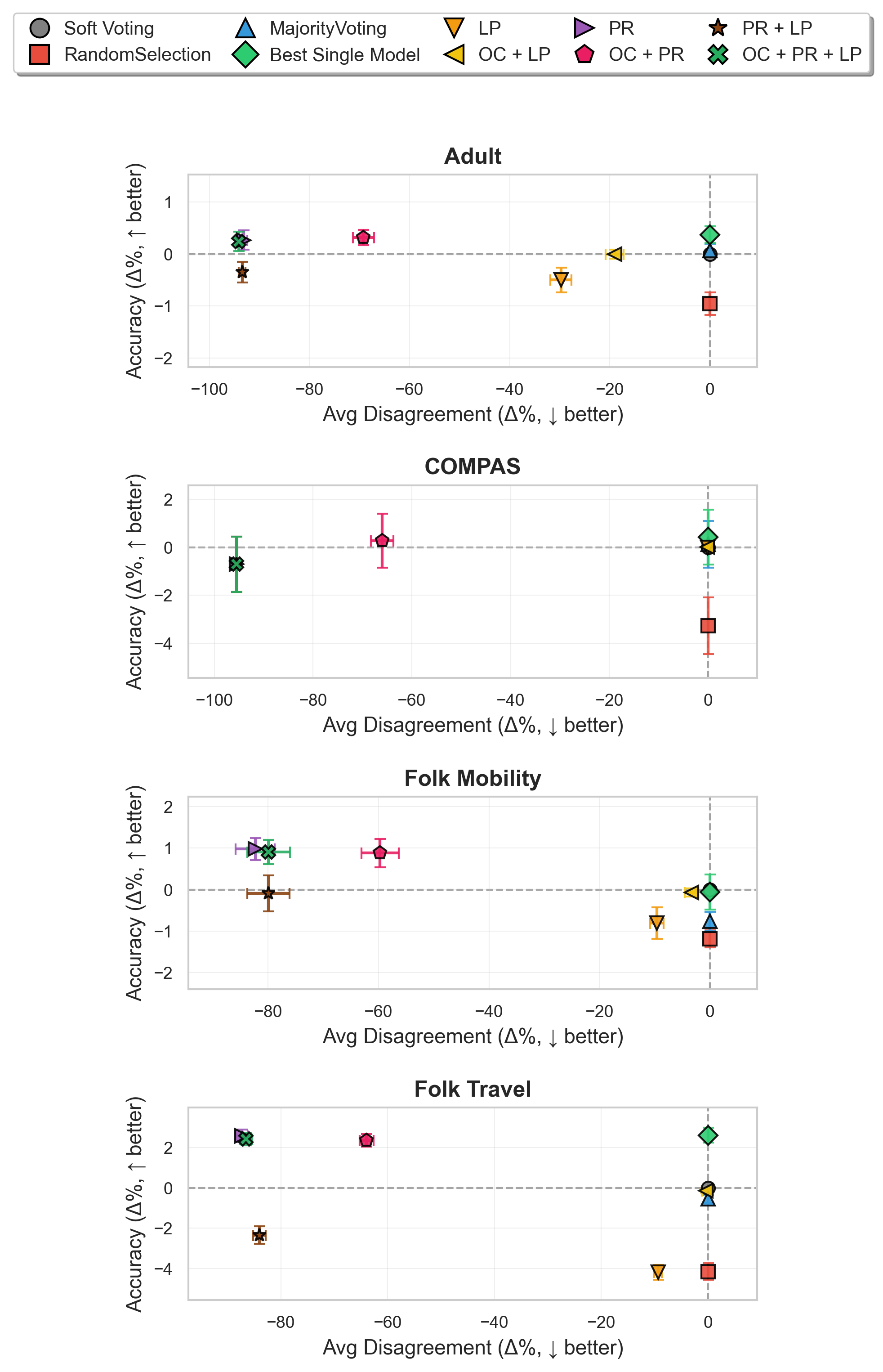}
  \vspace{-0.4em}
  \caption{Accuracy-disagreement trade-offs (mean $\pm$ std over 10 seeds; $\Delta\%$ vs.\ Soft Voting).}
  \label{fig:appendix_acc_disag}
  \vspace{-0.6em}
\end{figure}

\small
Figure~\ref{fig:appendix_acc_disag} visualizes the trade-off between predictive accuracy
($\Delta\%\mathrm{Acc}$, $\uparrow$) and ensemble disagreement
($\Delta\%\mathrm{Disag}$, $\downarrow$), both relative to Soft Voting.
PR drives disagreement close to zero; PR+LP and OC+PR+LP achieve similarly large reductions
with stable (often improved) accuracy, while LP alone can incur modest accuracy losses.
\normalsize

\subsection{Neighborhood Agreement Across Methods}
\begin{figure}[H]
  \centering
  \includegraphics[width=\columnwidth]{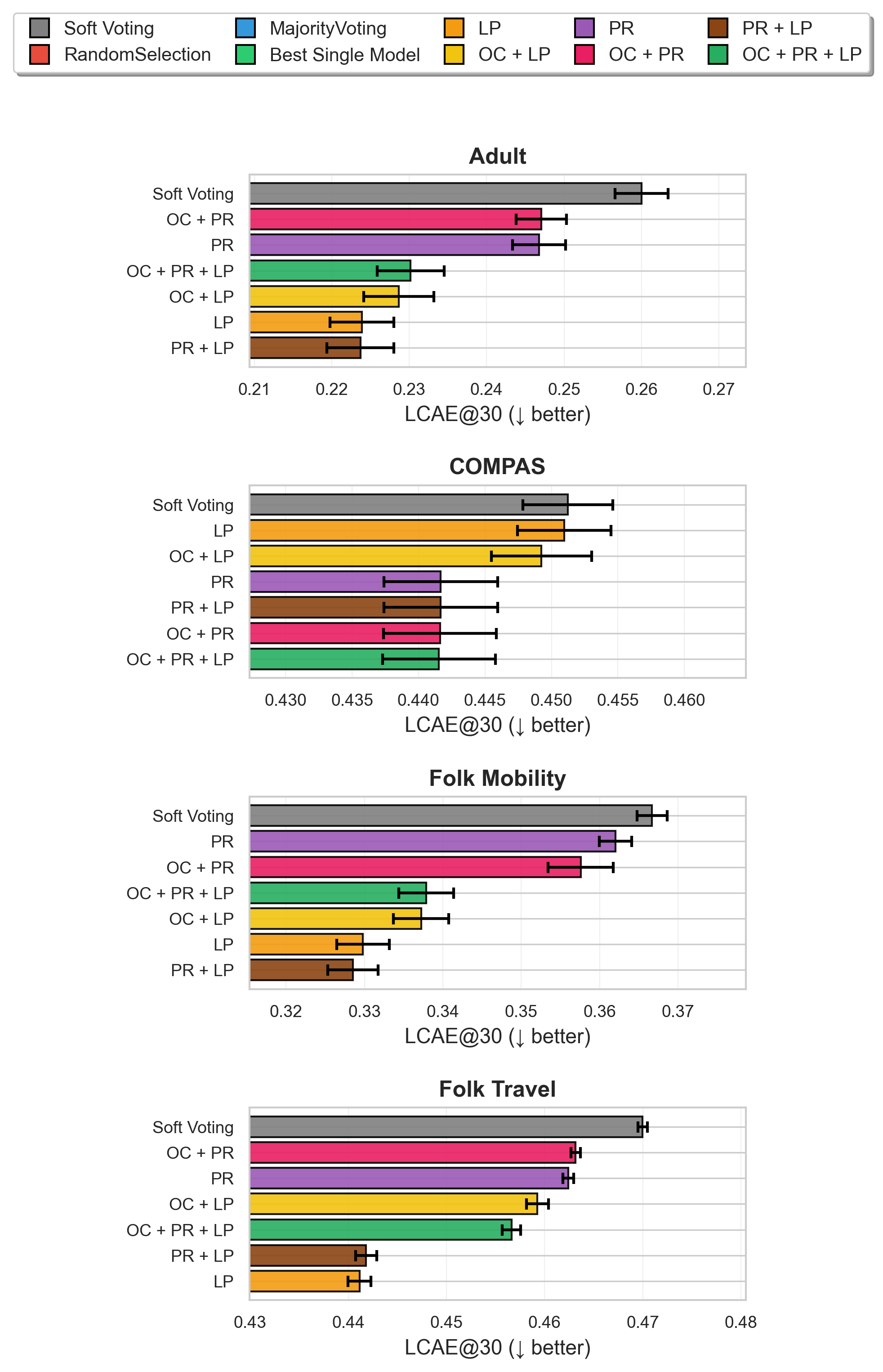}
  \vspace{-0.4em}
  \caption{Neighborhood agreement via $\mathrm{LCAE@30}$ (mean $\pm$ std over 10 seeds).}
  \label{fig:appendix_lcae}
  \vspace{-0.6em}
\end{figure}

\small
Figure~\ref{fig:appendix_lcae} reports neighborhood agreement via $\mathrm{LCAE@30}$ ($\downarrow$).
LP is the primary driver of LCAE improvements, and combining LP with PR yields the strongest overall
trade-offs (low disagreement with improved neighborhood agreement).
\normalsize

\subsection{Hyperparameter Settings}

\paragraph{Outlier Correction (OC).}
Outlier Correction uses two tunable hyperparameters:
the training flip rate $\rho_{\mathrm{train}}$ (fraction of training labels flipped)
and the validation soft-label rate $\rho_{\mathrm{val}}$
(fraction of validation labels replaced by the Rashomon ensemble mean).
Hyperparameters are selected via validation search to reduce downstream ensemble disagreement
subject to preserving predictive accuracy.
We tune
$\rho_{\mathrm{train}} \in \{0.005, 0.01, 0.02, 0.03, 0.05, 0.07, 0.10\}$
and
$\rho_{\mathrm{val}} \in \{0.002, 0.005, 0.01, 0.015, 0.02, 0.03\}$.
Across datasets, selected values were
$\rho_{\mathrm{train}} = 0.02$ and $\rho_{\mathrm{val}} \in \{0.01, 0.03\}$.
All other design choices are fixed: outliers are ranked by the absolute deviation between
the label and the Rashomon ensemble mean, training outliers are corrected via hard label flipping,
validation outliers receive soft labels equal to the ensemble mean, and models are retrained
on the corrected labels.

\paragraph{Local Patching (LP).}
Local Patching has two tunable hyperparameters:
the neighborhood size $k$ and the directional bias threshold $\tau_{\mathrm{bias}}$,
which controls the minimum fraction of one-sided residuals required to trigger a patch.
Hyperparameters are selected via validation search to improve neighborhood agreement
subject to the verification constraint that patches must not increase neighborhood Brier loss.
We tune
$k \in \{3,5,7,10,15,20,30,40,50,80,120,160,220,300\}$
and
$\tau_{\mathrm{bias}} \in \{0.51,0.55,0.60,0.65,0.70,0.75,0.80,0.85,0.90\}$.
Across datasets, small neighborhoods were selected in practice
($k=5$ with $\tau_{\mathrm{bias}}=0.6$).

\paragraph{Pairwise Reconciliation (PR).}
Pairwise Reconciliation uses the following tunable hyperparameters:
the signed disagreement threshold $\varepsilon$,
batch size $B$ (number of model pairs reconciled per iteration),
minimum disagreement region size $\alpha$,
convex weight $\lambda$ in the correction objective,
minimum required Brier improvement $\delta$,
stopping tolerance $\eta$,
and maximum number of iterations $T_{\max}$.
Hyperparameters are selected via validation to reduce ensemble disagreement
subject to the verification constraint that each accepted correction must improve
Brier loss on the corresponding disagreement region.
Across datasets, values include
$\varepsilon\in\{0.03,0.05\}$,
$B\in\{10,40\}$,
$\alpha=15$,
$\lambda=0.5$,
$\delta=10^{-4}$,
$\eta\in\{10^{-3},10^{-2}\}$,
and $T_{\max}\in\{10,200\}$.

\end{document}